\documentclass[conference]{IEEEtran}
\IEEEoverridecommandlockouts
\usepackage{cite}
\usepackage{amsmath,amssymb,amsfonts}
\usepackage{algorithmic}
\usepackage{graphicx}

\usepackage{textcomp}
\usepackage{hyperref}

\usepackage{booktabs, multirow} 
\usepackage{soul}
\usepackage[table]{xcolor} 
\usepackage{changepage,threeparttable} 
\usepackage{url}
\let\OLDthebibliography\thebibliography
\renewcommand\thebibliography[1]{
  \OLDthebibliography{#1}
  \setlength{\parskip}{0pt}
  \setlength{\itemsep}{0pt plus 0.3ex}
}
\usepackage{enumitem}

\newcommand{\etal}{\textit{et al.}}

\newcommand{\eg}{\textit{e.g.}}

\begin{document}

\title{Exposing Lip-syncing Deepfakes from \\Mouth Inconsistencies
}

\author{\IEEEauthorblockN{Soumyya Kanti Datta, Shan Jia, Siwei Lyu}
\IEEEauthorblockA{\textit{University at Buffalo, State University of New York} \\
\{soumyyak, shanjia, siweilyu\}@buffalo.edu}
}

\maketitle

\begin{abstract}
A lip-syncing deepfake is a digitally manipulated video in which a person's lip movements are created convincingly using AI models to match altered or entirely new audio. Lip-syncing deepfakes are a dangerous type of deepfakes as the artifacts are limited to the lip region and more difficult to discern. 
In this paper, we describe a novel approach, \underline{LIP}-syncing detection based on mouth \underline{INC}onsistency (LIPINC), for lip-syncing deepfake detection by identifying temporal inconsistencies in the mouth region. These inconsistencies are seen in the adjacent frames and throughout the video. Our model can successfully capture these irregularities and outperforms the state-of-the-art methods on several benchmark deepfake datasets. Code is available at \url{https://github.com/skrantidatta/LIPINC}.
\end{abstract}


\section{Introduction}

\label{sec:intro}


Deepfakes are digitally altered videos in which a person's appearance and voice are modified to mimic someone else, often for realistic impersonations, leveraging advanced generative AI methods. Various forms of Deepfakes include face swapping \cite{faceswap}, face reenactment\cite{Faceforensics++}, lip-syncing\cite{lipsync}, and face animation\cite{zeng2022fnevr}. While there are positive uses of deepfakes, such as in movie dubbing, reviving historical figures, and creating virtual avatars for individuals with speech impairments by synthesizing voices similar to their own, deepfakes pose significant societal risks. 
They can be misused for identity theft, fabricating fake news, or generating fraudulent videos for purposes like extortion or spreading disinformation. Accessible open-source tools\footnote{\eg, Wav2Lip: \url{https://github.com/Rudrabha/Wav2Lip},\\ DeepFaceLab: \url{https://github.com/iperov/DeepFaceLab}, \\ FakeApp: \url{https://www.malavida.com/en/soft/fakeapp/}.} allow the creation of deepfake videos featuring celebrities or public figures with minimal technical expertise.
A notable recent instance is the circulation of deepfake videos on social media, falsely depicting CNN and Fox News anchors endorsing novel or simplistic diabetes treatments\footnote{\url{https://today.rtl.lu/news/fact-check/a/2136220.html}}.

\begin{figure}[ht]
\centering
\includegraphics[width=\columnwidth]{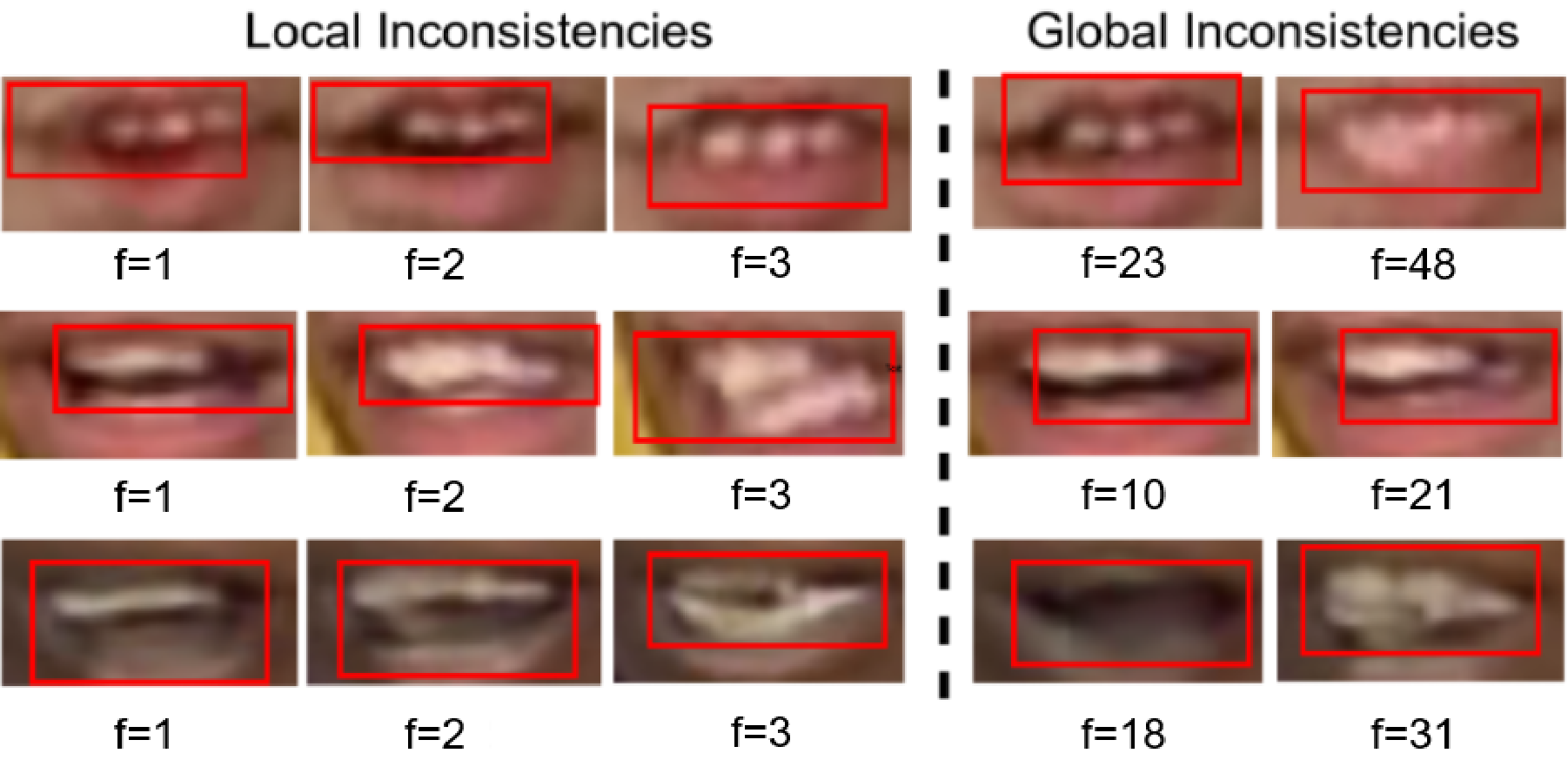}
\caption{\em \small Illustration of the mouth inconsistency in lip-syncing deepfakes. We visualize video frames from FakeAVCeleb~\cite{FakeAVCeleb} dataset with open mouths. Here f represents frame number. The first three columns present consecutive frames for local comparison, while the last two columns offer a broader perspective by displaying frames with similar poses from the entire video, defined as global inconsistencies in our paper. The deepfakes exhibit more pronounced inconsistencies in aspects such as mouth shape, coloration, dental structure, and tongue appearance.} 
\label{fig:example}
\end{figure}

Deepfakes are broadly categorized into two types based on the regions of manipulation: entire-face synthesis and partial manipulation. Entire-face synthesis encompasses techniques such as face-swapping and talking head generation. On the other hand, partial manipulation primarily involves lip-syncing videos. In these videos, the lip movements of an individual are altered to match a specific audio track, creating the illusion that they are speaking words they originally did not. Lip-synced deepfakes focus on modifying only a person's mouth area, which facilitates the creation of high-quality deepfakes. A recent investigation \cite{sundstrom2023deepfake} indicates that lip-synced deepfakes pose a greater challenge for human detection than face-swapping deepfakes. However, unlike the extensively researched face-swap deepfake detection~\cite{FTCN,li2018ictu,li2018exposing,nguyen2022deep,muppalla2023integrating}, identifying lip-syncing deepfakes presents greater challenges and is less explored.

\begin{figure*}[t]
\centering
\includegraphics[width=0.85\textwidth]{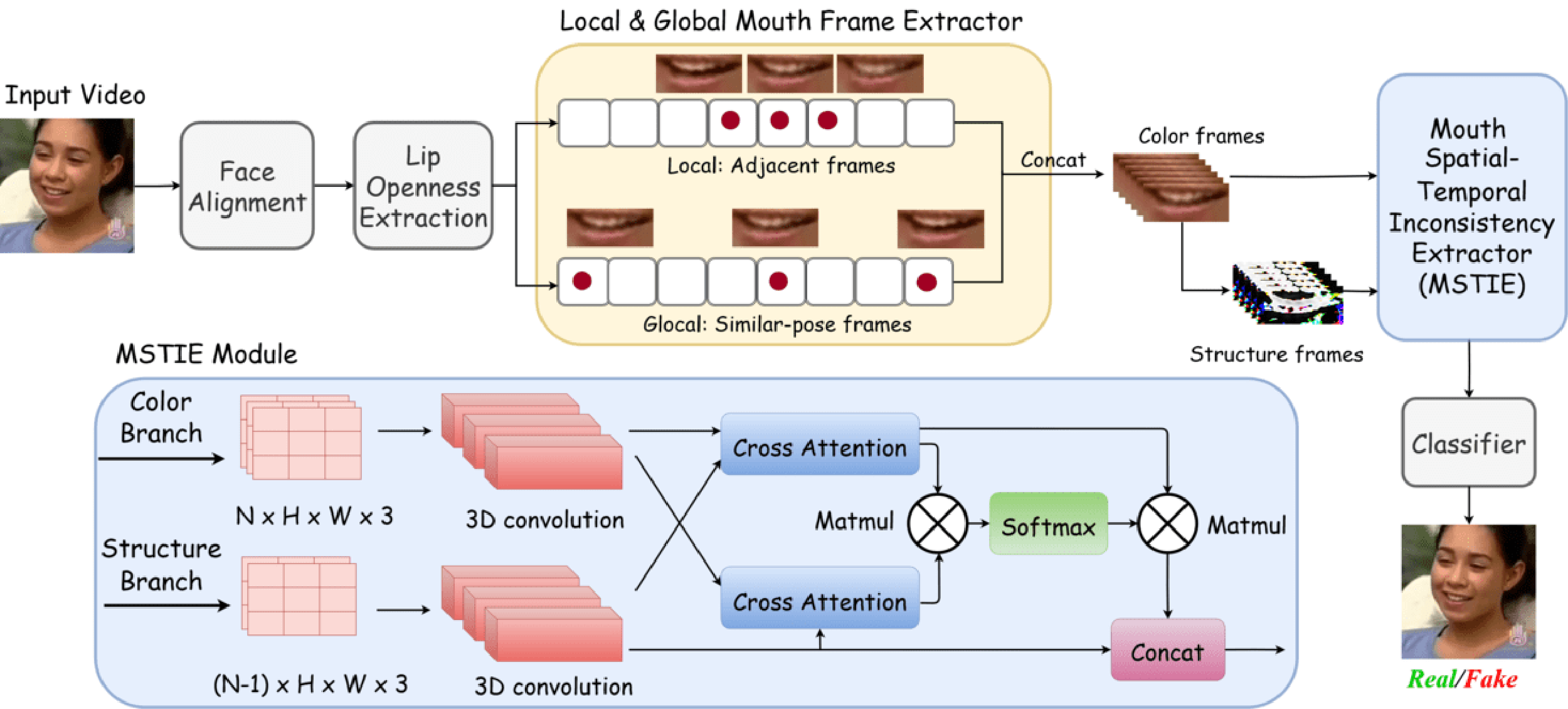}
\caption{\em \small End to End pipeline of the proposed LIPINC model.}
\label{fig:LIPINC}
\end{figure*}


In recent developments, lip-syncing detection methodologies have focused on learning features related to speech-audio inconsistency and dis-synchronization \cite{agarwal2023watch, feng_et_al}. Diverging from the reliance on motion-based~\cite{Lips-don't-lie}, frame-level~\cite{tolosana2021deepfakes}, or synchronization-focused features, our work introduces the LIPINC model, which investigates spatial-temporal inconsistency patterns to distinguish between lip-synced and authentic videos. The following observations inspire this approach: in real videos, open mouths in successive frames typically show similar characteristics, a principle we term `local consistency'. Similarly, mouths in akin poses across different video segments should exhibit uniformity, referred to as `global consistency'. However, in lip-synced deepfakes, maintaining such consistency in mouth details across both local and global contexts is challenging. As illustrated with examples from the FakeAVCeleb dataset~\cite{FakeAVCeleb} in Fig. \ref{fig:example}, lip-synced videos reveal noticeable discrepancies in mouth shape, color, and the appearance of teeth and tongue. While these spatial and temporal irregularities are often not obvious to the human eye in real-time playback, our method aims to systematically capture these unique patterns to improve the accuracy of deepfake detection. 
Our work has three contributions. 
\begin{itemize}[leftmargin=1em] \itemsep 0em
 \item Our analysis reveals that spatial-temporal inconsistencies in the mouth region across adjacent and non-adjacent frames with similar mouth poses are telltale signs of lip-syncing deepfake videos.
\item We have developed the LIPINC pipeline, which comprises a module for locally and globally matching mouth poses, and a feature extractor for inconsistencies, all underpinned by a novel inconsistency loss function.
\item Our experimental evaluations on three distinct lip-syncing datasets demonstrate the LIPINC method's exceptional ability to detect both in-domain and previously unseen lip-syncing deepfakes. 
\end{itemize}


\section{Related Work}


\subsection{DeepFake Detection}
Most deepfake detection approaches focus on entire-face synthesis in videos, identifying visual anomalies like face warping~\cite{li2018exposing}, facial movements~\cite{trinh2021interpretable}, discrepancies between face and context~\cite{nirkin2021deepfake}, and inconsistencies in face identity~\cite{huang2023implicit}. These methods extract distinctive features from the entire face and are less effective when dealing with partially manipulated lip-syncing videos. For lip-sync deepfake detection, various techniques are employed to exploit audio-visual mismatches. Shahzad \etal \cite{LipSyncMatters} proposed a method to detect fabricated videos by contrasting the real lip sequence from a video with a synthetic lip sequence created based on audio using a lip-sync model \cite{lipsync}. Feng \etal \cite{feng_et_al} investigated audio-visual signal inconsistencies in forged videos through anomaly detection, but this approach might miss manipulations that achieve good synchronization~\cite{feng_et_al}. Alternatively, Haliassos \etal \cite{Lips-don't-lie} identified forged videos by detecting semantic anomalies in mouth movements from video segments, focusing on irregularities in mouth motion from lower face regions, which may be less effective in videos with minimal mouth movement. LipForensics~\cite{Lips-don't-lie}, in contrast, learned mouth motion patterns using lipreading networks from lower-face clips for deepfake identification. Tolosana \etal \cite{tolosana2021deepfakes} examined visual artifacts across various facial regions (eyes, nose, and mouth), utilizing spatial indicators to differentiate between real and fake videos.

\subsection{Lip-syncing Deepfake Datasets}
Early deepfake datasets are for entire-face synthesis, such as face-swapping and face-reenactment~\cite{Faceforensics++,Celeb-df}. The emergence of Wav2Lip~\cite{lipsync} in 2020 marked a significant advancement in lip-syncing deepfakes, and led to the development of several datasets aimed at detecting partially manipulated lip-syncing videos. For instance, the FakeAVCeleb \cite{FakeAVCeleb} dataset is a comprehensive audio-video deepfake detection resource featuring accurately lip-synced videos. This dataset encompasses three categories of audio-video deepfakes sourced from English-language YouTube videos, including lip-syncing \cite{lipsync}, face-swapping \cite{faceswap,Fsgan}, and voice cloning \cite{voice-clones}. Additionally, the KODF \cite{Kodf} dataset presents a vast collection of both generated and authentic videos with Korean subjects, with the fake videos produced using Wav2Lip and various face-swapping models. The CMDFD~\cite{yu2024explicit} dataset covers two lip-syncing models and two talking
head generation models for cross-modal Deepfake detection. 
Note that these existing datasets encompass only a subset of lip-syncing videos created with a single model. 

\section{Method}

The diversity in video attributes such as resolution, head poses, facial expressions, and lighting conditions poses significant challenges for developing an effective detection model. To address these complexities, we introduce the LIPINC model to identify spatial-temporal mouth inconsistencies on both local and global scales for deepfake detection. The model evaluates a video to ascertain its authenticity, categorizing each video with a class label $c \in \{0,1\} $, where 0 denotes Fake, and 1 indicates Real.

The overall pipeline of the proposed LIPINC model is depicted in Fig. \ref{fig:LIPINC}. It comprises two primary modules: 1) \textbf{Local and Global Mouth Frame Extractor}, which isolates adjacent and similarly posed mouth segments based on mouth openness throughout the video sequence; and 2) \textbf{Mouth Spatial-Temporal Inconsistency Extractor} (MSTIE), responsible for encoding and learning distinctive inconsistency features within and across frames, focusing on both color and structure aspects. The output of the MSTIE module is then directed to a binary classifier to determine the probabilities of each class.

\subsection{Local and Global Mouth Frame Extractor}
\begin{figure}[h]
\centering
\includegraphics[width=0.23\textwidth]{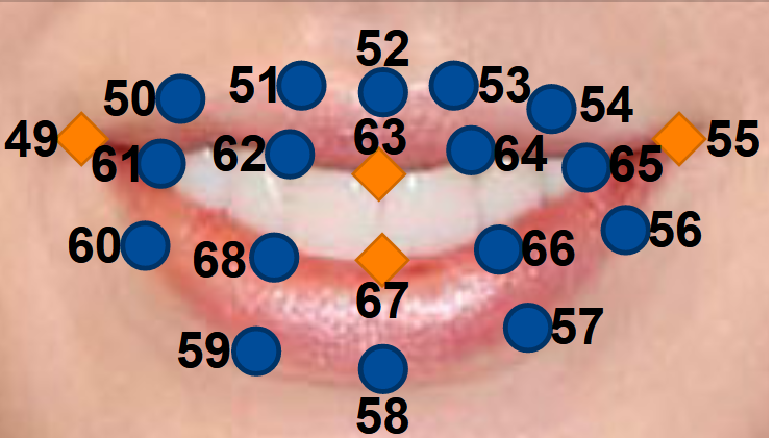}
\caption{\em \small Mouth region landmarks detected by Dlib \cite{Dlib}. Orange colors denote the landmarks for mouth openness measurement and matching.}
\label{fig:dlib}
\end{figure}

 Given an input video, we first use a face detector, such as Dlib~\cite{Dlib}, to crop and align the face in each video. 
 Then we use the facial landmarks to extract the mouth region, as shown in Fig. \ref{fig:dlib}. 
 We propose to extract multiple frames with open mouths since they contain more inconsistencies regarding deepfakes. For the adjacent mouth extraction as local inconsistency analysis, we obtain a sequence of $L$ adjacent frames with open mouth, which are selected based on a large distance between landmarks 63 and 67 in Fig. \ref{fig:dlib}. 
 For the global inconsistency representation, we search for $G$ similar pose frames identical to the mouth openness in local frames from the rest of the video. {The similar pose frames are extracted by comparing the height and width of opened-lips with similar face orientations}. To ensure a global matching and prevent selected frames from being adjacent, we set the extraction of similar-pose frames to have a minimum time gap of 0.09 seconds between them. We then concatenate the adjacent and similar-pose frames to form our color sequence $\textbf{C}$ with $N$ frames where $\textbf{C}\in R^{(N*H*W*3)}$, 3 is the channel number, and $N=L+G$. 
 To emphasize the consistency description at the mouth structure level, we further add the structure branch with residual sequence $\textbf{S}$, where $\textbf{S} \in R^{((N-1)*H*W*3)}$ is calculated by
 \begin{equation}
      S_t = C_{t+1} - C_t \; (1\le t<N)
 \end{equation}
{The structural branch is dedicated to capturing the inconsistencies in mouth and teeth shapes across different frames.}
The color and structure sequences with local and global mouth frames are fed into the following inconsistency feature extraction module.

\begin{table*}[ht]\centering
\centering
\begin{tabular}{l|r|rrrr|rr|rrr}\toprule
\multirow{2}{*}{Method} &\multirow{2}{*}{Source} &\multicolumn{4}{|c|}{FakeAVCeleb (FakeAV-LS)} &\multicolumn{2}{c|}{LSR+W2L} &\multicolumn{2}{c}{KODF-LS} \\\cmidrule{3-10}
& &Precision &Accuracy &AP &AUC &AP &AUC &AP &AUC \\\midrule
LipSyncMatters \cite{LipSyncMatters} &\cite{LipSyncMatters} &0.96 &0.97 &- &- &- &- &- &- \\
Feng et al. \cite{feng_et_al} &\cite{feng_et_al} &- &- &0.94 &0.94 &- &- &0.88 &0.87 \\
Intra-modal \cite{tian2023unsupervised} & \cite{tian2023unsupervised}&- &- &0.94 &0.67 &- &- &- &- \\
Intra-Cross-modal \cite{tian2023unsupervised} & \cite{tian2023unsupervised}&- &- &\textbf{0.99} &0.96 &- &- &- &- \\
RealForensics \cite{RealForensics} &\cite{feng_et_al} &- &- &0.91 &0.95 &- &- &\textbf{0.96} &\textbf{0.94} \\
FTCN \cite{FTCN} &\cite{feng_et_al} &- &- &0.96 &\textbf{0.97} &- &- &0.67 &0.68 \\
\midrule
LipForensics \cite{Lips-don't-lie}  & RI &0.89 & 0.81 &0.93 &0.95 &0.83 &0.86 &0.88 &0.87 \\
Xception \cite{Faceforensics++} & RI &0.83 &0.80 &0.87 &0.85 &0.72 &0.79 &0.73 &0.75 \\
\midrule
LIPINC (ours) & RI &\textbf{0.99} &\textbf{0.99} &0.94 &\textbf{0.97} &\textbf{0.87} & \textbf{0.87} &0.90 &0.91 \\
\bottomrule
\end{tabular}
\vspace{0.15cm}
\caption{  \textbf{In-domain testing on FakeAVCeleb dataset \cite{FakeAVCeleb} and cross-domain testing on KODF \cite{Kodf} and LSR+W2L}. We compare our model with state-of-the-art detectors regarding Precision, Accuracy, Average Precision and AUC score. Here RI means Re-Implemented. Note that the results sourced from existing methods used much more training data than us for in-domain cross-type testing.}\label{tab:1}
\end{table*}

 \subsection{Mouth Spatial-Temporal Inconsistency Extractor}
This MSTIE module is designed to encode the color and structure frames and learn discriminative features representing spatial and temporal inconsistency levels for deepfake detection. We use the 3D-CNN \cite{3D_CNN} model to produce the spatial-temporal features $\textbf{v}_{ci}$ and $\textbf{v}_{si}$ for the color and the structure sequence, respectively, where $i$ is the number of input video. 
 These vectors are then passed through a cross-attention \cite{vaswani2017attention} module which interconnects the color and structure branches such that $K_{ci}= V_{ci} = \textbf{v}_{ci},$ $Q_{ci} = \textbf{v}_{si}$ and  $K_{si}= V_{si} = \textbf{v}_{si}$, $Q_{si} = \textbf{v}_{ci}$, where $K,V$ and $Q$ represent the keys, values, and queries. The cross attention for color $A_c$ and structure $A_s$ branch is calculated as follows:
\begin{equation}
 A_{ci} = \text{softmax}\left(\frac{Q_{ci}K_{ci}^T}{\sqrt{d_{qc}}}\right)V_{ci}
\end{equation} 

\begin{equation}
A_{si} = \text{softmax}\left(\frac{Q_{si}K_{si}^T}{\sqrt{d_{qs}}}\right)V_{si}
\end{equation} 

Next, the output of the color branch $A_{ci}$ and structure branch $A_{si}$ are fused together using a cross-attention module while focusing more weight on the structure branch by concatenating with the structure features, as through ablation analysis we found that structure features produces better generalization performance.
\begin{equation}
 A_{fi} = \text{softmax}\left(\frac{Q_{i}K_{i}^T}{\sqrt{d_Q}}\right)V_{i} + \textbf{v}_{si}
\end{equation} 
Here $K_{i}= V_{i} = A_{ci},$ $Q_{i} = A_{si}$ and here $d_q$ and $d_Q$ represents the feature dimensions of the respective queries. The output is subsequently fed into a binary classifier for deepfakes prediction.


\subsection{Loss Function}
We design an inconsistency loss, $L_{IL}$, to measure the inconsistency level between the video frames. Given a set of input mouth frames, we feed it to our MSTIE module and extract the feature vector $F$ from the first layer of the 3D-CNN module where $F \in R^{(N'*H'*W'*C')}$  where $N', H', W'$, and $C'$ are frame number, height, width, and channel size, respectively. For each frame in $F$, we compare it against all other frames in $F$ and calculate the pairwise similarity using Structural Similarity Index (SSIM)\cite{SSIM}. The similarity scores range from 0 to 1, where 1 indicates the frames are similar and vice versa. We then extract the average similarity score to get $AvgS$. For a real video we expect the average similarity score to be close to 1 while for fake videos, it should be near zero. We use a cross-entropy loss to supervise the average similarity score. Therefore, the inconsistency loss is calculated by:
\begin{equation}
 L_{IL} = - \frac{1}{M}\sum_{i=1}^{M}y_ilog(AvgS_i) + (1-y_i)log(1-AvgS_i)
\end{equation}
where $M$ is the number of input videos, $y_i \in \{0,1\}$ where $y_i=0$ denotes a fake video and $y_i=1$ denotes a real video.

\begin{figure}[t]
\centering
\includegraphics[height=0.73\columnwidth]{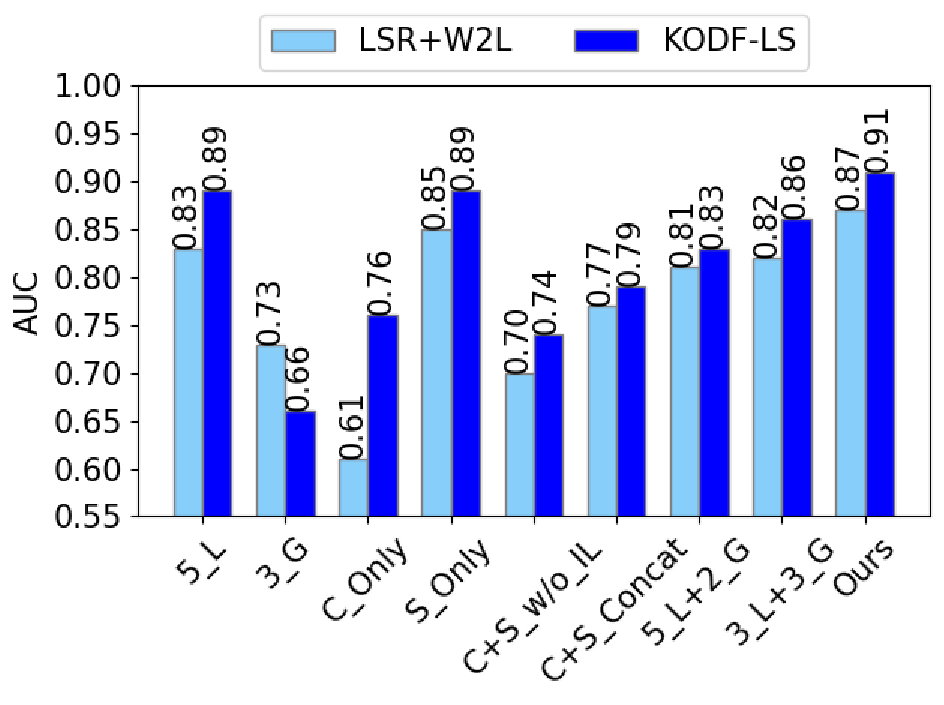}
\caption{\em \small Ablation analysis on LSR+W2L and KODF-LS Dataset based on AUC scores. Here  L, G, C, and S refer to Local, Global, Color, and Structure frames, respectively. IL is Inconsistency Loss.}
\label{fig:ablation}
\end{figure}

In addition to the inconsistency loss, we also use the classification loss $L_{CL}$ to optimize the feature learning, which is a binary cross-entropy loss. Thus, the total loss function of our method is $L_{total} = \lambda_1 L_{CL} + \lambda_2 L_{IL}$, where $\lambda_1$ is the classification loss weight, and $\lambda_2$ is the inconsistency loss weight. Ablation studies (\ref{fig:ablation}) show that $\lambda_1=1$ and $\lambda_2=5$ give the best overall performance.

\section{Experiments}

\subsection{Settings}
\noindent\textbf{Datasets.} 
We use the FakeAVCeleb dataset\cite{FakeAVCeleb} to train and validate our model. The FakeAVCeleb dataset consists of 500 real unique frontal face videos and more than 20,000 multi-modal deepfake videos. 
We split the dataset into two categories: Lip-synced (FakeAV-LS) deepfakes generated by the Wav2lip~\cite{lipsync} model and Face-swap (FakeAV-FS) deepfakes that have not used Wav2lip. Since our method mainly focuses on lip-syncing deepfakes detection, we have divided the lip-synced deepfakes in this dataset into a non-overlapped training set and a test set. We use two more datasets to evaluate our model's generalization ability. The first one is the KODF dataset~\cite{Kodf}, which includes 62,166 real videos and 175,776 deepfake videos. Similar to the FakeAVCeleb dataset, deepfake videos contain lip-syncing and entire-face synthesis. We created 3 test sets from the dataset based on the deepfake generation method. 1) KODF-LS: audio-driven lip-synced deepfakes from \cite{lipsync, ATFHP}; 2) KODF-FSGAN: Faceswap deepfakes generated using \cite{Fsgan}; 3) KODF-DFL:Faceswap deepfakes generated using \cite{DeepFaceLab-paper}. We randomly selected a subset of real and fake videos for evaluation following previous studies~\cite{feng_et_al}, and use 800 samples for each class. Secondly, we have generated a new dataset called LSR+W2L, which is specifically tailored for lip-syncing detection. We consider two models, Wav2Lip and Wav2Lip with GAN model~\cite{lipsync}. This dataset contains 1,000 real videos from the LRS2 dataset \cite{lsr2} and 1,000 fake videos. 

\noindent\textbf{Evaluation Metrics.} The proposed LIPINC model is evaluated using four common metrics in deepfake detection, including Precision, Accuracy, Average Precision (AP), and Area Under the Curve (AUC) scores \cite{AUC}. 

\noindent\textbf{Experimental Settings.} In the proposed architecture, we use the Dlib toolkit \cite{Dlib} to extract $64\times144$ mouth regions from the input video. We set the local frame number $L$ as five and the global frame number $G$ as three such that the model can detect the required number of global frames even if the video is only a few seconds long. For all experiments, we use Adam optimizer\cite{adam} with 0.001 learning rate and 0.1 epsilon. We train our model for 100 epochs with a batch size 16 on the Keras framework with TensorFlow version 2.15.0.

\subsection{Comparison under In-domain Testing}
In Table \ref{tab:1}, we first evaluate our model on the FakeAVCeleb testing set (FakeAV-LS) \cite{FakeAVCeleb} and compared its performance with the state-of-the-Art (SOTA) methods. The proposed LIPINC model achieves the best performance in terms of accuracy, precision and is tied with the FTCN~\cite{FTCN} model for the best performance in terms of AUC scores. Our model outperforms the best SOTA method Intra-Cross-modal~\cite{tian2023unsupervised} in terms of AUC scores by 1\% and performs slightly worse than it in AP. Overall, the LIPINC model demonstrates outstanding performance in detecting lip-syncing videos under in-domain testing, thanks to the utilization of spatial-temporal mouth inconsistency features.  



\subsection{Comparison under Cross-domain Testing}
We further evaluate our model on KODF-LS\cite{Kodf} and LSR+W2L datasets to evaluate the generalization capability of our model. The results are shown in Table \ref{tab:1}. Note that our model was trained on the FakeAVCeleb training set (FakeAV-LS), while the SOTA results cited from existing papers were trained on the whole FakeAVCeleb dataset \cite{FakeAVCeleb} as introduced in \cite{feng_et_al}. On our newly created LSR+W2L dataset, the proposed LIPINC model outperforms all the baselines in terms of AP and AUC scores. The RealForensics model achieves better performance than our method on the KODF dataset due to its utilization of a large training dataset. Our method achieves an AUC of over 87\% on both datasets, demonstrating strong generalization in detecting unknown lip-syncing videos.

\subsection{Cross-type Deepfake Testing}

Furthermore, to demonstrate the performance of our LIPINC model across different types of deepfake videos, such as faceswap deepfakes, we evaluate its performance on three face-swap datasets using the model trained on the FakeAV-LS set. The LIPINC model achieves an AUC score of 91\% on the FakeAV-FS dataset as in-domain cross-type testing, while an AUC of 60\% and 67\% on the KODF-FSGAN and KODF-DFL \cite{Kodf} datasets, respectively. The drop in performance on the KODF dataset suggests the presence of distinct mouth inconsistency patterns between lip-syncing videos and face-swap deepfakes.



\subsection{Ablation Studies}
In this section, we conduct ablation studies to validate the efficacy of different modules in our model on the two datasets, namely LSR+W2L and KODF-LS \cite{Kodf}. Our proposed LIPINC model utilizes five local frames, three global frames, color and structure branches in the MSTIE module, and an inconsistency loss to detect deepfakes. 
We consider several variants of our detection model, including C\_Only which refers to the model with only the color branch without the MSTIE module, S\_Only which refers to the model with only a structure branch, C+S\_Concat indicating a model with two branches without the MSTIE module, and C+S\_w/o\_IL indicating a model without our inconsistency loss. 
We use L for the local frames and G for the global frames. The associated number refers to the number of frames for that case. 
The comparison results are shown in Fig. \ref{fig:ablation}. It clearly demonstrates the contribution of each module in our model architecture and how variations in a number of local or global frames influence the performance of our model. 


Specifically, it can be observed that the removal of any module results in an obvious decrease in the generalization ability of the model. First, removing all the local frames (3\_G) affects the performance the most, where the AUC decreases by 14\% and 25\% on the LSR+W2L and KODF\_LS, respectively, whereas the removal of the global frames (5\_L) reduced the AUC by 4\% and 2\% on two datasets respectively. Second, removing the inconsistency loss (C+S\_w/o\_IL) decreased the AUC by 17\% for both datasets. Similarly, when we delete the MSTIE module, the performance of C\_Only decreases by 26\% and 15\%, respectively. However, it should be noted that the performance of S\_Only remains adequate, with the AUC decreasing by just 2\% for both LSR+W2L and KODF-LS, showing the importance of structure features in inconsistency learning. Interestingly, when the color branch is added back to the S\_Only model, namely the C+S\_Concat strategy, the performance deteriorates. This happens because using the concatenate layer instead of our MSTIE to fuse the outputs of the color and structure branch cannot effectively capture the inconsistencies. Last, we find that decreasing the number of the total frames slightly affects the generalization performance. It can be noted that the reduction of global frames from 3 to 2 (5\_L+2\_G)  affected the performance more than the reduction of local frames from 5 to 3 (3\_L+3\_G).

\subsection{Limitations}
Although our model performs well in detecting lip-syncing deepfakes, our approach cannot detect deepfake images and videos lacking lip movement, or videos that are too short to detect global frames.

\section{Conclusion}

In this paper, we have proposed a novel method called LIPINC to detect lip-syncing deepfakes. We demonstrate that the mouth region in deepfake videos contains spatial-temporal discrepancies. These inconsistencies are obvious in the consecutive frames and throughout the video. Our model can capture these inconsistency patterns by designing a novel mouth spatial-temporal inconsistency extractor guided by an inconsistency loss. Extensive experimental results on three datasets show the outstanding performance of our method in detecting both in-domain and cross-domain deepfakes. Our future work will explore incorporating the audio modality to enhance lip-syncing detection performance.

\noindent\textbf{Acknowledgement}. This work is supported in part by the US Defense Advanced Research Projects Agency (DARPA) Semantic Forensic (SemaFor) program, under Contract No. HR001120C0123, National Science Foundation (NSF) Projects under grants SaTC-2153112, No.1822190, and TIP-2137871, and University at Buffalo's Office of Vice President for Research and Economic Development and Center for Information Integrity. The views and conclusions contained herein are those of the authors and should not be interpreted as necessarily representing the official policies, either expressed or implied, of DARPA, NSF, or the U.S. Government.


{
\bibliographystyle{IEEEbib}
\bibliography{icme}
}
\end{document}